# A Full Transformer-based Framework for Automatic Pain Estimation using Videos

Stefanos Gkikas[1] and Manolis Tsiknakis[2]

*Abstract*— The automatic estimation of pain is essential in designing an optimal pain management system offering reliable assessment and reducing the suffering of patients. In this study, we present a novel full transformer-based framework consisting of a Transformer in Transformer (TNT) model and a Transformer leveraging cross-attention and self-attention blocks. Elaborating on videos from the *BioVid* database, we demonstrate state-of-the-art performances, showing the efficacy, efficiency, and generalization capability across all the primary pain estimation tasks.

## I. INTRODUCTION

Pain, according to Williams and Craig [1], is "a distressing experience associated with actual or potential tissue damage with sensory, emotional, cognitive and social components". From a biological perspective, pain is an unfavorable sensation that originates from the peripheral nervous system. Its primary function is to activate sensory neurons and alert the organism to potentially harmful situations, thus serving as a vital mechanism for identifying and responding to threats [2]. The Global Burden of Disease (GBD) study refers that pain is the number one cause of years lived with disability (YLD), concerning not only individuals but also society as a whole, constituting clinical, economic, and social constraints [3]. The primary types of pain are acute and chronic. The major difference between them is related to the duration; when it is present for less than three months, the pain is considered acute and probably accompanied by physical damage, while chronic perseveres the recovery process [4]. People of all ages experience painful situations due to an accident, illness, or even during treatment, provoking a plethora of daily life challenges. Especially in chronic pain conditions, additional mental health problems, *e.g.*, anxiety, depression, and sleep-related problems, commonly occur [5]. Furthermore, inadequate pain management often leads to negative collateral consequences associated with drug overuse, opioids, and addiction [6]. A crucial matter that needs focused attention is the welfare of vulnerable groups who may not be able to communicate directly or objectively. Their pain assessment is usually based on observing behavioral or physiological responses from caregivers or family members. This specific setting often leads to wrong or insufficient assessment for two main reasons: continuous monitoring is challenging without adopting technology-based solutions, and the estimation's precision is often minimal due to inadequate training or prejudices [7]. Further challenges arise with the elderly where either diminished manifestation ability or even unwilling communication behavior are presented [8]. In addition, an essential body of research [9][10] indicates significant variations of pain manifestation among people of different gender and age, suggesting that the pain assessment is an even more intricate process requiring increased consideration. The automatic pain estimation procedure is founded on utilizing behavioral and physiological modalities. The primary behavioral modalities include facial expressions, body-head movements, gestures, and vocalizations, while the physiological include electrocardiography, electromyography, and skin conductance responses.

The mainstream deep neural architectures in computer vision (CV) are the Convolutional Neural Networks (CNN). Especially in the research field of automatic pain assessment elaborating images/videos, CNNs are the fundamental component of every approach [5]. The domination of transformer architecture [11] in natural language processing (NLP), where their core element is the self-attention mechanism, inspired researchers to develop equivalent models for visual applications. The introduction of Vision Transformers (ViT) [12] led to the creation of a new paradigm of architecture in the computer vision domain. A plethora of new approaches has developed on the basis of ViT. Such an approach is the Transformer in Transformer (TNT) [13], which enhances the local feature representation by the further division of the patches into sub-patches. Despite the impressive results and flexibility of the transformer-based models, they scale poorly with the input size and increase the computational cost because of the self-attention layers which compare the input to every other input. Several efforts have been made to reduce the complexity and improve the efficiency of such architectures. The primary approach is the replacement of self-attention with cross-attention [14] or the incorporation of both [15].

In this study, we develop a framework consisting of a TNT model, which is utilized as the *"spatial feature extraction module"* applied to each video frame, and a transformer-based model with cross and self-attention blocks as the *"temporal feature extraction module"* applied to each feature sequence of videos. In this way, we can exploit the temporal dimension of videos and offer more reliable estimations about the continuous nature of the pain sensation. The remaining of this study is organized as follows: in Section

[1]Stefanos Gkikas is a Ph.D. candidate in Affective Computing at Hellenic Mediterranean University, Heraklion, Greece `gkikas@ics.forth.gr`

[2]Manolis Tsiknakis is a Professor of Biomedical Informatics at Hellenic Mediterranean University and Affiliated Researcher at the Computational Biomedicine Laboratory of the Foundation for Research and Technology (FORTH), Heraklion, Greece `tsiknaki@ics.forth.gr`

II, we present the related work, and Section III describes the development process of our framework. Section IV presents the conducted experiments and results, and finally, Section V concludes the paper.

## II. RELATED WORK

Numerous research efforts have been made to estimate the pain level in humans utilizing videos. Zhi et Wan [16] trying to capture pain's dynamic nature, they developed Long Short-term Memory Networks with sparse coding (SLSTM). Tavakolian *et al.* [17] developed 3D CNNs with kernels of various temporal depths capturing short, mid, and long-range facial expressions. Similarly, in [18], the authors proposed a 3D CNN but combined it with self-attention structures to increase the importance of specific input dimensions. Thiam *et al.* [19] adopted two strategies to exploit video's temporal dimension. Initially, they encoded the video frames into motion history and optical flow images, and after, they designed a framework incorporating a CNN and a bidirectional LSTM (biLSTM). In a similar manner, the authors in [20] also encoded videos into single RGB images employing statistical spatio-temporal distillation (SSD) and followed by a Siamese network trained in a self-supervised setting. Werner *et al.* [21] followed a domain-specific feature approach, proposing a set of markers describing facial actions and their dynamics and classifying them with a deep random forest (RF) classifier, while Patania *et al.* [22] adopted Deep Graph Neural Networks (GNN) architectures and dense maps of fiducial points in order to detect pain. Finally, Xin *et al.* [23] presented a multi-task framework, estimating the person's identity beyond the pain level, comprising a CNN with an autoencoder attention module.

Regarding transformer-based methods, the only study proposed in [24] where the authors developed a deep attention transformer framework that consists of a ResNet subnetwork extracting frame-based features and a transformer model capturing the temporal relationship among the frames.

## III. METHODOLOGY

This section describes the employed database, the preprocessing methods, the design of our framework, as well as implementation details regarding the training procedure.

### A. Preprocessing

Before feeding videos into our model for the pain estimation procedure, it was necessary to apply face detection and alignment for performance and computational efficiency improvements. We combined the well-known face detector MTCNN [25] with the Face Alignment Network (FAN) [26], which utilizes 3D landmarks. The 3D approach is essential to our problem since the head movements in several cases, especially in high-intensity pain, are increased, leading to erroneous alignment from 2D approaches. We also note that all the experiments were conducted utilizing frames of resolution $224 \times 224$ pixels. Figure 1 depicts the facial alignment method applied in a video frame.

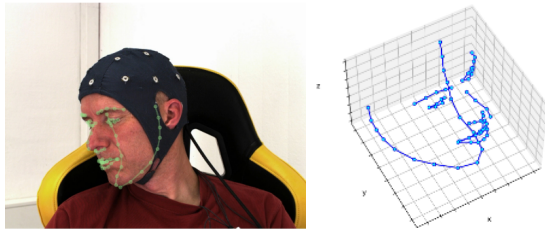

Fig. 1: Application of the face alignment. Illustration of the landmarks in 2D space (left) and 3D space (right).

### B. Transformer-based Framework

Our framework consists of two main components; the *"spatial feature extraction module"*, *i.e.*, a TNT model, and the *"temporal feature extraction module"*, *i.e.*, a transformer with cross and self-attention blocks. In Figure 2, we illustrate our proposed framework, which consists of 24 million parameters and 4.2 giga floating point operations (GFLOPS).

*1) Spatial feature extraction module:* Similarly to standard ViT, every given frame is initially split into $n$ patches $\mathcal{F}^k = [F^{k,1}, F^{k,2}, ... F^{k,n}] \in \mathbb{R}^{n \times p \times p \times 3}$, where $p \times p$ is the resolution of each patch (*i.e.*, $16 \times 16$) and 3 is the number of color channels. Afterward, the patches are further divided into $m$ sub-patches for the model to learn both global and local feature representations of the frame. Consequently, every input frame of a video is transformed into a sequence of patches and sub-patches:

$$\mathcal{F}^k \rightarrow [F^{k,n,1}, F^{k,n,2} ..., F^{k,n,m}], \quad (1)$$

where $F^{k,n,m} \in \mathbb{R}^{s \times s \times 3}$ is the $m$-th sub-patch of $n$-th patch of $k$-th frame of each video, while $s \times s$ is resolution of each sub-patch (*i.e.*, $4 \times 4$). Next, the patches and the sub-patches with a linear projection are transformed into embeddings $Z$ and $Y$. The following step is the position embedding, where the spatial information of each patch and sub-patch is retained. This procedure is based on the 1D learnable position encoding, where for each patch, the following position encodings is assigned:

$$Z_0 \leftarrow Z_0 + E_{patch}, \quad (2)$$

where $E_{patch}$ are the patch position encodings. Respectively, for each sub-patch within a patch, a position encoding is added:

$$Y_0^i \leftarrow Y_0^i + E_{sub-patch}, \quad (3)$$

where $E_{sub-patch}$ are the sub-patch position encodings and $i = 1, 2, ... m$ is the index of a sub-patch within a patch. Next, the sub-patches are led to a transformer encoder called an "Inner Transformer Encoder", consisting of 2 multi-head self-attention blocks, which are essentially dot product attention. The attention is expressed as follows:

$$Attention(Q, K, V) = softmax\left(\frac{QK^T}{\sqrt{d_k}}V\right), \quad (4)$$

where $Q \in \mathbb{R}^{M \times D}$, $K \in \mathbb{R}^{M \times C}$ and $V \in \mathbb{R}^{M \times C}$ ($M$ is input dimension, $C$ and $D$ are channel dimensions) are projections of the corresponding input and represent the Query, Key and Value matrices respectively. Specifically, $Q = XW_Q$, $K = XW_K$ and $V = XW_V$ where $W$ are the learnable weight matrices and $X$ is the input. The output embedding of the "Inner Transformer Encoder" is added to the patch embedding, which subsequently is led to the "Outter Transformer Encoder". This encoder consists of 3 multi-head self-attention blocks, and the output embedding of it is a feature vector $d = 192$. The entire *"spatial feature extraction module"* has a depth of 12 blocks.

*2) Temporal feature extraction module:* The extracted feature embedding of each frame within a video is concatenated into a larger vector $\mathcal{D}$ which essentially is a feature representation of the entire video $\mathcal{V} \Rightarrow \mathcal{D} = (d^1 \frown d^2 \frown, ...d^k)$. Afterward, the D feature vector is driven into our temporal module, a transformer model consisting of 1 cross-attention and 2 self-attention components with a fully connected neural network (FCN) after each one. The cross-attention introducing asymmetry into the attention operation reduces the computational complexity and makes our approach more efficient. Specifically, instead of a projection of the input with dimensions of $M \times D$, the $Q$ in cross-attention is a learned matrix with dimensions $N \times D$, where $N < M$. The self-attention components of this module are identical as described in equation 4. The cross and self-attention blocks are 1 and 8 multi-heads, respectively. Furthermore, regarding position encoding, we adopted the Fourier feature position encoding [15].

*3) Training Details:* Initially, before the automatic pain estimation training procedure, we pre-trained our *"spatial feature extraction module"* with the *VGGFace2* dataset [27], consisting of more than three million face images from over nine thousand people. In Table I, we list the hyper-parameters of our method and the augmentation techniques applied.

*C. Database Details*

In this study, we employed the publicly available *BioVid Heat Pain Database* [28], which incorporates facial videos, electrocardiograms, electromyograms, and skin conductance levels of 87 healthy participants (subjects). They were subjected to experimentally induced heat pain at five different intensity levels; No pain (NP), mild pain ($P_1$), moderate pain ($P_2$), severe pain ($P_3$), and very severe pain ($P_4$). The participants were stimulated 20 times for each intensity, thus generating 100 samples for every modality. In this work, we utilized the videos ($87 \times 100 = 8700$) from Part A of *BioVid*.

## IV. EXPERIMENTS & RESULTS

In this section, we present the conducted experiments regarding pain estimation. We note that the experiments were performed in binary and multi-level classification settings. Specifically, (1) NP vs. $P_1$, (2) NP vs. $P_2$, (3) NP vs. $P_3$, (4) NP vs. $P_4$ respecting the binary classification tasks, and finally, (5) multi-level pain classification, utilizing all the available pain classes of the database. The evaluation protocol that we followed is the leave-one-subject-out (LOSO) cross-validation. Furthermore, the classification metrics adopted in this study are the following: micro-average accuracy, macro-average precision, macro-average recall (sensitivity), and macro-average F1 score.

TABLE I: Training hyper-parameters used in our method.

| Epochs | Optimizer | Learning rate | LR decay | Weight decay | Warmup epochs |
|---|---|---|---|---|---|
| 200 | AdamW | 1e-4 | cosine | 0.1 | 5 |
| Label smoothing | DropPath | Attention Dropout | Loss Function | Augmentation Methods | |
| 0.1 | 0.1 | 0.1 | Cross Entropy | *AugMix* & *TrivialAugment* | |

DropPath applied to the *"spatial feature extraction module"*, Attention Dropout applied to the *"temporal feature extraction module"*

*A. Pain Estimation*

Regarding the classification results of the pain estimation tasks, we observe the following: on NP vs. $P_1$, we achieved $65.95\%$ accuracy, while the precision is close to it with $65.90\%$. Similarly, the F1 score is $65.04$, and interestingly the recall (sensitivity) is $67.85\%$. On NP vs. $P_2$, the accuracy increased to $66.87\%$ as also the other performance metrics, especially the F1 score, which increased over $1.15\%$ showing the improvement in the detection of true positive samples. On NP vs. $P_3$, the increase in the performances is particularly noticeable. We attained $69.22\%$ accuracy, while the sensitivity improved to $70.84\%$. The classification improvement is reasonable since the pain is characterized as severe at the $P_3$ level, and the subjects' manifestations become more intense. On the task with the higher level of pain, *i.e.,* NP vs. $P_4$, the recall is $74.75\%$, while in terms of accuracy, we achieved $73.28\%$. It is evident that recognizing very severe pain is the most straightforward identification task considering that the pain threshold is on the tolerance limits, and most subjects demonstrate it clearly with their facial expressions. Finally, the range of performances is diminished in the last task, *i.e.,* the multi-level classification, since estimating all levels simultaneously is a more challenging procedure. We attained $31.52\%$ accuracy and recall of $29.94\%$, indicating that the ability to detect true positive samples in this task has more challenges.

At this point, we want to highlight that our framework regarding both the architecture and the training procedure remained identical across all tasks, binary and multi-level classification tasks. Our purpose was to study the generalization capabilities of our method for every possible scenario (within the limits of the database) similar to clinical settings. Table II presents the classification results.

*B. Video Sampling*

In this section, we study the effect of video sampling on automatic pain estimation. The experiments in IV-A were conducted utilizing all the available frames (*i.e.,* 138) from each video. In the following experiments, we sample frames with a stride of 2, 3, and 4. Initially, utilizing all 138 frames leads to a video feature representation $\mathcal{D}$ with a size of

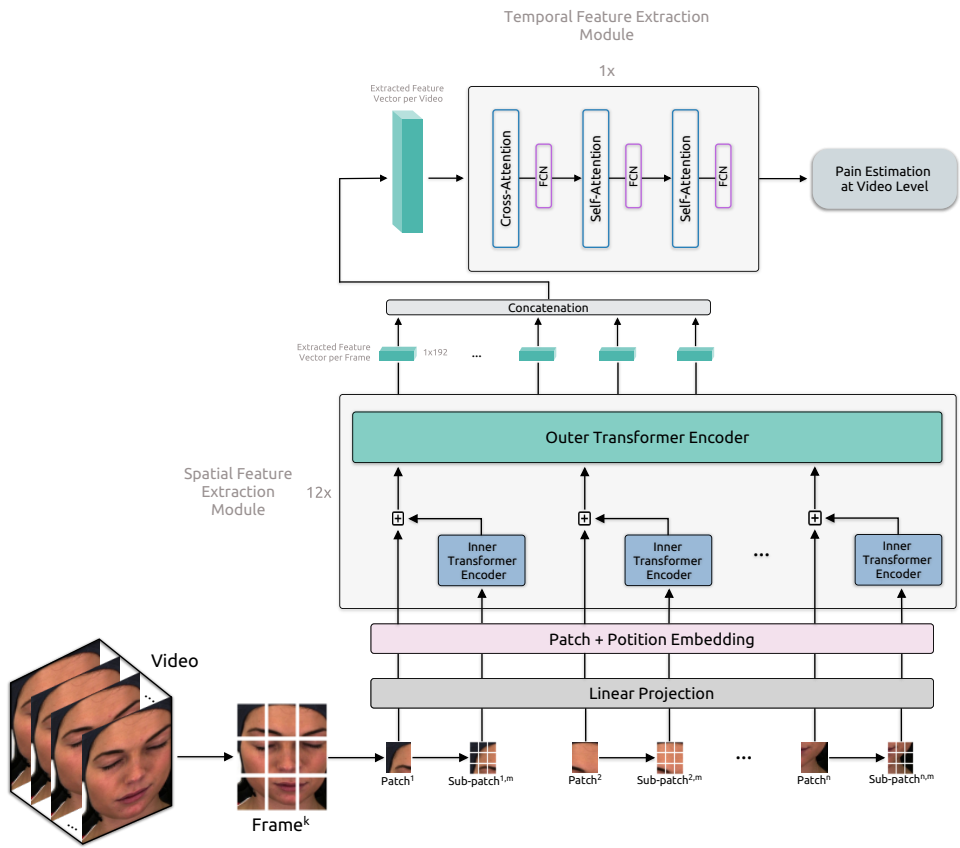

Fig. 2: An overview of our proposed framework for automatic pain estimation

TABLE II: Classification results on the pain estimation tasks.

| Metric | Task | | | | |
|---|---|---|---|---|---|
| | NP vs $P_1$ | NP vs $P_2$ | NP vs $P_3$ | NP vs $P_4$ | MC |
| Acc. | 65.95 | 66.87 | 69.22 | 73.28 | 31.52 |
| Pre. | 65.90 | 66.89 | 69.18 | 73.31 | 31.48 |
| Rec. | 67.85 | 68.34 | 70.84 | 74.75 | 29.94 |
| F1 | 65.04 | 66.19 | 68.54 | 72.75 | 27.82 |

Acc.: accuracy  Pre.: precision  Rec.: recall  NP: no pain  $P_1$: mild pain  $P_2$: moderate pain  $P_3$: severe pain  $P_4$: very severe pain  MC: multi-level classification $138 \times 192 = 26496$. Likewise, a stride of 2 leads to 69 frames, and a size of $\mathcal{D}$ equals $69 \times 192 = 13248$. For strides 3 and 4, we have 46 and 35 frames, respectively, and sizes of $\mathcal{D}$ equal 8832 and 6720. Table III presents the classification accuracy using the different number of input frames for the corresponding pain estimation tasks. At the same time, Figure 3 illustrates the impact of the number of frames on the mean accuracy across the five tasks and the mean runtime during inference. We observe an increase in performance of about 1.38%, utilizing 138 frames compared to 35 frames. Respectively, the runtime is increased by a factor of 3. Despite the fact of the multiply of runtime, every sampling rate choice can achieve real-time automatic pain estimation in situations where it is needed.

TABLE III: Classification results on the pain estimation tasks utilizing a different number of input frames, reported on accuracy %.

| Number of Frames | Task | | | | |
|---|---|---|---|---|---|
| | NP vs $P_1$ | NP vs $P_2$ | NP vs $P_3$ | NP vs $P_4$ | MC |
| 138 | 65.95 | 66.87 | 69.22 | 73.28 | 31.52 |
| 69 | 65.76 | 66.74 | 69.15 | 73.25 | 31.29 |
| 46 | 65.66 | 66.70 | 68.50 | 71.78 | 31.20 |
| 35 | 65.40 | 66.12 | 68.32 | 72.01 | 30.80 |

### C. Interpretation

An important area of research, especially in the deep learning-related fields, is the interpretability of the models to provide explanations for the decisions making. This is especially true regarding healthcare topics since the transparency improvement of these models is essential for their acceptance and adoption in the clinical domain. In this study, we adopted the method of [29] to create relevance maps displaying in which facial areas our model, *i.e.,* the *"spatial feature extraction module"*, pays attention. Examples of the relevance maps are shown in Figure 4. We notice that in the initiation of a facial expression sequence, the model attends in "arbitrary" areas. As the pain progression continues, the attention becomes more precise to regions that manifest the painful occurrence. We want to point out that according to the relevance maps, no universal expressions describe pain exclusively. Nevertheless, we recognize a tendency toward general facial regions such as the mouth and eyes.

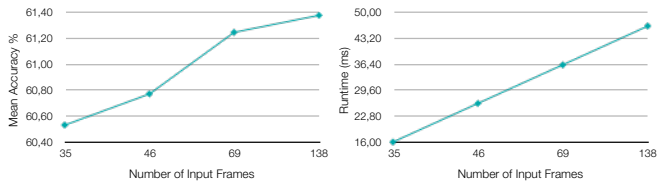

Fig. 3: The effect of the number of input frames on the accuracy (left) and the effect on the runtime in milliseconds (right). Runtime is during inference on a Nvidia RTX-3090.

*D. Comparison with existing methods*

Finally, in this section, we compare our accomplished results employing the transformer-based framework (using all available frames per video) with studies that utilize the Part A of the *BioVid* database with all $87$ subjects and follow the identical evaluation protocol, *i.e.,* leave-one-subject-out (LOSO) cross-validation, in order to perform objective and accurate comparisons. Table IV shows the corresponding results, where there are three main groups of studies; *i)* studies conducting exclusive pain detection (NP vs. $P_4$), *ii)* studies examing pain detection and multi-level pain estimation, and finally, *iii)* studies exploring all the major pain-related tasks.

Our approach, comparing it with the studies that conducted experiments on every task, achieved the highest performances on both binary and multi-level pain estimations. Regarding the studies that were performed solely on pain detection or/and multi-level pain estimation, our method attained comparable or even better results, *e.g.,* [19][20][22]. We observe that the performances come from the restricted in terms of experiments studies tend to be higher. In our view, however, the importance of researching and developing systems capable of performing adequately in every scenario is greater.

## V. CONCLUSIONS

This study explored the application of the transformer-based architecture for automatic pain estimation using videos. We developed a framework that consisted exclusively of transformer models, exploiting both the spatial and temporal dimensions of the frame sequences. The conducted experiments revealed the efficacy of our framework in assessing pain and demonstrating the generalization capabilities to accomplish every pain estimation task with satisfactory classification results, especially in low-intensity pain where the facial expressions are subtle. Furthermore, we showed that our proposed framework is characterized by high efficiency and is able to perform in real-time settings. Another important aspect of our study is the creation of relevance maps demonstrating to which facial areas the model pays attention. We believe that more efforts from the affective-computing community are needed to improve the interpretability of the adopted deep-learning approaches.

In addition, we suggest that future studies include details regarding the computational cost of their approaches, *e.g.,* throughput measurements, number of model parameters, or number of FLOPS, to assess their real-time application. Although it may not be the primary focus of the studies, it is still relevant. The comparison with other related methods demonstrated comparable or improved results depending on the corresponding training scenario. We believe that future research regarding the automatic pain estimation field needs to investigate all available tasks since they clinically provide essential information for pain management.

It is worth noting that our current approach may benefit from the utilization of more complex modules. Specifically, increasing the number of attention heads and blocks in the transformer models, or enlarging the extracted feature vectors, would result in a more comprehensive representation of the data. However, it should be acknowledged that implementing such modifications would also come with a significant increase in computational cost and time requirements.

## ACKNOWLEDGMENT

This work has received funding from the European Union's Horizon 2020 research and innovation program under grant agreement No $945175$ (Project: *CARDIOCARE*). This paper reflects only the author's view and the Commission is not responsible for any use that may be made of the information it contains.

## REFERENCES

[1] A. C. de C Williams and K. D. Craig, "Updating the definition of pain.," *Pain*, vol. 157, pp. 2420–2423, 11 2016.

[2] K. S and T. RS, "Neuroanatomy and neuropsychology of pain," *Cureus*, vol. 9, 10 2017.

[3] "Global, regional, and national incidence, prevalence, and years lived with disability for 354 Diseases and Injuries for 195 countries and territories, 1990-2017: A systematic analysis for the Global Burden of Disease Study 2017," *The Lancet*, vol. 392, pp. 1789–1858, nov 2018.

[4] D. C. Turk and R. Melzack, "The measurement of pain and the assessment of people experiencing pain.," pp. 3–16, 2011.

[5] S. Gkikas and M. Tsiknakis, "Automatic assessment of pain based on deep learning methods: A systematic review," *Computer Methods and Programs in Biomedicine*, vol. 231, p. 107365, 2023.

[6] P. Dinakar and A. M. Stillman, "Pathogenesis of Pain," *Seminars in Pediatric Neurology*, vol. 23, pp. 201–208, aug 2016.

[7] B. G. S. Dekel, A. Gori, A. Vasarri, M. C. Sorella, G. Di Nino, and R. M. Melotti, "Medical evidence influence on inpatients and nurses pain ratings agreement," *Pain Research and Management*, vol. 2016, 2016.

[8] H. H. Yong, S. J. Gibson, D. J. Horne, and R. D. Helme, "Development of a pain attitudes questionnaire to assess stoicism and cautiousness for possible age differences.," *The journals of gerontology. Series B, Psychological sciences and social sciences*, vol. 56, pp. P279–84, sep 2001.

[9] E. J. Bartley and R. B. Fillingim, "Sex differences in pain: a brief review of clinical and experimental findings," *British journal of anaesthesia*, vol. 111, pp. 52–58, jul 2013.

[10] S. Gkikas., C. Chatzaki., E. Pavlidou., F. Verigou., K. Kalkanis., and M. Tsiknakis., "Automatic pain intensity estimation based on electrocardiogram and demographic factors," pp. 155–162, SciTePress, 2022.

[11] A. Vaswani, N. Shazeer, N. Parmar, J. Uszkoreit, L. Jones, A. N. Gomez, L. Kaiser, and I. Polosukhin, "Attention is all you need," in *Proceedings of the 31st International Conference on Neural Information Processing Systems*, NIPS'17, (Red Hook, NY, USA), p. 5998–6008, Curran Associates Inc., 2017.

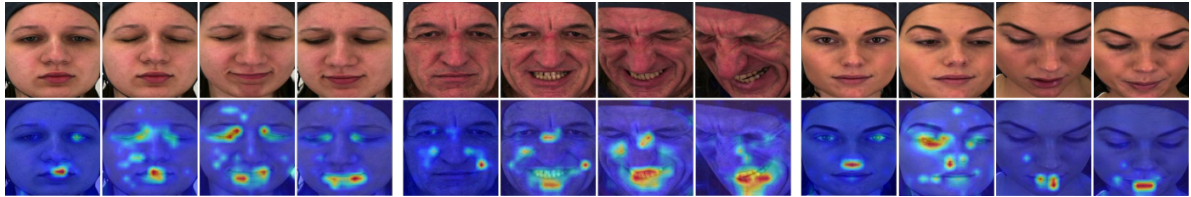

Fig. 4: Relevance Maps

TABLE IV: Comparison of studies that utilized BioVid, videos, and LOSO cross-validation

| Study | Task | | | | |
|---|---|---|---|---|---|
| | NP vs $P_1$ | NP vs $P_2$ | NP vs $P_3$ | NP vs $P_4$ | MC |
| Tavakolian & Hadid. [17] | - | - | - | 86.02 | - |
| Thiam et al. [19] | - | - | - | 69.25 | - |
| Tavakolian et al. [20] | - | - | - | 71.02 | - |
| Patania et al. [22] | - | - | - | 73.20 | - |
| Huang et al. [18] | - | - | - | 77.50 | 34.30 |
| Xin et al. [23] | - | - | - | 86.65 | 40.40 |
| Zhi & Wan [16] | 56.50 | 57.10 | 59.60 | 61.70 | 29.70 |
| Werner et al. [21] | 53.30 | 56.00 | 64.00 | 72.40 | 30.80 |
| Our approach | 65.95 | 66.87 | 69.22 | 73.28 | 31.52 |

[12] A. Dosovitskiy, L. Beyer, A. Kolesnikov, D. Weissenborn, X. Zhai, T. Unterthiner, M. Dehghani, M. Minderer, G. Heigold, S. Gelly, *et al.*, "An image is worth 16x16 words: Transformers for image recognition at scale," *arXiv preprint arXiv:2010.11929*, 2020.
[13] K. Han, A. Xiao, E. Wu, J. Guo, C. Xu, and Y. Wang, "Transformer in transformer," *Advances in Neural Information Processing Systems*, vol. 34, pp. 15908–15919, 2021.
[14] J. Lee, Y. Lee, J. Kim, A. Kosiorek, S. Choi, and Y. W. Teh, "Set transformer: A framework for attention-based permutation-invariant neural networks," in *Proceedings of the 36th International Conference on Machine Learning* (K. Chaudhuri and R. Salakhutdinov, eds.), vol. 97 of *Proceedings of Machine Learning Research*, pp. 3744–3753, PMLR, 09–15 Jun 2019.
[15] A. Jaegle, F. Gimeno, A. Brock, O. Vinyals, A. Zisserman, and J. Carreira, "Perceiver: General perception with iterative attention," in *International conference on machine learning*, pp. 4651–4664, PMLR, 2021.
[16] R. Zhi and M. Wan, "Dynamic facial expression feature learning based on sparse RNN," in *Proceedings of 2019 IEEE 8th Joint International Information Technology and Artificial Intelligence Conference, ITAIC 2019*, pp. 1373–1377, Institute of Electrical and Electronics Engineers Inc., may 2019.
[17] M. Tavakolian and A. Hadid, "A Spatiotemporal Convolutional Neural Network for Automatic Pain Intensity Estimation from Facial Dynamics," *International Journal of Computer Vision*, vol. 127, pp. 1413–1425, oct 2019.
[18] D. Huang, X. Feng, H. Zhang, Z. Yu, J. Peng, G. Zhao, and Z. Xia, "Spatio-temporal pain estimation network with measuring pseudo heart rate gain," *IEEE Transactions on Multimedia*, vol. 24, pp. 3300–3313, 2022.
[19] P. Thiam, H. A. Kestler, and F. Schwenker, "Two-stream attention network for pain recognition from video sequences," *Sensors (Switzerland)*, vol. 20, p. 839, feb 2020.
[20] M. Tavakolian, M. Bordallo Lopez, and L. Liu, "Self-supervised pain intensity estimation from facial videos via statistical spatiotemporal distillation," *Pattern Recognition Letters*, vol. 140, pp. 26–33, 2020.
[21] P. Werner, A. Al-Hamadi, K. Limbrecht-Ecklundt, S. Walter, S. Gruss, and H. C. Traue, "Automatic pain assessment with facial activity descriptors," *IEEE Transactions on Affective Computing*, vol. 8, no. 3, pp. 286–299, 2016.
[22] S. Patania, G. Boccignone, S. Buršić, A. D'Amelio, and R. Lanzarotti, "Deep graph neural network for video-based facial pain expression assessment," in *Proceedings of the 37th ACM/SIGAPP Symposium on Applied Computing*, pp. 585–591, 2022.
[23] X. Xin, X. Li, S. Yang, X. Lin, and X. Zheng, "Pain expression assessment based on a locality and identity aware network," *IET Image Processing*, vol. 15, no. 12, pp. 2948–2958, 2021.
[24] H. Xu and M. Liu, "A deep attention transformer network for pain estimation with facial expression video," in *Chinese Conference on Biometric Recognition*, pp. 112–119, Springer, 2021.
[25] K. Zhang, Z. Zhang, Z. Li, and Y. Qiao, "Joint face detection and alignment using multitask cascaded convolutional networks," *IEEE signal processing letters*, vol. 23, no. 10, pp. 1499–1503, 2016.
[26] A. Bulat and G. Tzimiropoulos, "How far are we from solving the 2d & 3d face alignment problem?(and a dataset of 230,000 3d facial landmarks)," in *Proceedings of the IEEE International Conference on Computer Vision*, pp. 1021–1030, 2017.
[27] Q. Cao, L. Shen, W. Xie, O. M. Parkhi, and A. Zisserman, "Vggface2: A dataset for recognising faces across pose and age," in *2018 13th IEEE international conference on automatic face & gesture recognition (FG 2018)*, pp. 67–74, IEEE, 2018.
[28] S. Walter, S. Gruss, H. Ehleiter, J. Tan, H. C. Traue, S. Crawcour, P. Werner, A. Al-Hamadi, A. O. Andrade, and G. M. D. Silva, "The biovid heat pain database: Data for the advancement and systematic validation of an automated pain recognition," pp. 128–131, 2013.
[29] H. Chefer, S. Gur, and L. Wolf, "Transformer interpretability beyond attention visualization," in *Proceedings of the IEEE/CVF Conference on Computer Vision and Pattern Recognition*, pp. 782–791, 2021.